\documentclass[letterpaper, 10 pt, conference]{ieeeconf}  

\IEEEoverridecommandlockouts                              

\usepackage{tabularx} 
\usepackage{multicol}
\usepackage[bookmarks=true]{hyperref}
\usepackage{bm}
\usepackage{amsbsy}
\usepackage{amssymb}  
\usepackage{mathtools}

\usepackage{import}
\usepackage{pgfplots}
\usepackage{color}
\usepackage{tabularx} 
\usepackage{multirow}
\usepackage{pgf}

\usepackage{balance}

\newcommand\Tstrut{\rule{0pt}{.6ex}}         

\newcommand{\vect}[1]{\boldsymbol{#1}}
\newcommand{\norm}[1]{\left\lVert#1\right\rVert}

\newtheorem{definition}{Definition}[section]

\title{\LARGE \bf
ULT-model: Towards a one-legged unified locomotion template model for forward hopping with an upright trunk
}

\author{Dennis Ossadnik, Elisabeth Jensen and Sami Haddadin$^{*}$
	\thanks{$^{*}$All authors are with the Munich School of Robotics and Machine Intelligence,
		Technical University of Munich, 80797 Munich, Germany
		{\tt\small \{dennis.ossadnik, elisabeth.jensen, haddadin\}@tum.de}}%
}

\begin{document}

\maketitle
\thispagestyle{empty}
\pagestyle{empty}

\begin{abstract}

While many advancements have been made in the development of template models for describing upright-trunk locomotion, the majority of the effort has been focused on the stance phase. In this paper, we develop a new compact dynamic model as a first step toward a fully unified locomotion template model (ULT-model) of an upright-trunk forward hopping system, which will also require a unified control law in the next step. We demonstrate that all locomotion subfunctions are enabled by adding just a point foot mass and a parallel leg actuator to the well-known trunk SLIP model and that a stable limit cycle can be achieved. This brings us closer toward the ultimate goal of enabling closed-loop dynamics for anchor matching and thus achieving simple, efficient, robust and stable upright-trunk gait control, as observed in biological systems.

\end{abstract}

\section{INTRODUCTION}

Accurate model-based simulations of legged locomotion are useful both from a biological perspective, for deeper understanding of neuromuscular function, as well as from a technological perspective, for developing robust and effective control methods for legged robots. A great leap forward in the understanding of bipedal locomotion was the discovery that core locomotor subfunctions can be described by lower-dimensional neuromechanical models called template models, which take advantage of symmetries and synergies \cite{Full_99}. Template models may be embedded in morphological or physiological models, known as anchors, where more complex physics and controls are represented. While the templates enable high-level abstraction and simpler control mechanisms (which may reflect true neuromechanical phenomena in humans and animals), anchors more closely reflect the true physics of the system and thus reveal deeper causal explanations. A number of excellent template models have been proposed over the last thirty years, generally describing either the stance or the swing phase of gait. \par

\subsection{Background and state of the art} \label{sec:background}
The stance phase has been given significantly more attention in gait modeling than swing, because it has the greatest effect on stability and acceleration. The spring-loaded inverted pendulum (SLIP) model (first proposed by Blickhan \cite{Blickhan89}) and its variations is the most widely used model for the stance phase of gait. Some of the most common variations include the bipedal SLIP model \cite{Geyer2006} and the SLIP model in which the point mass is replaced with a rigid trunk with the hips offset from the center of mass (CoM), whose orientation is controlled according to a virtual pivot point (VPP-TSLIP) \cite{Maus2008}. Such trunk-based models are especially important for representing trunk stabilization, which is a significant challenge associated with legged locomotion. \par
\begin{figure}
	\centering
	\def\svgwidth{0.4\textwidth}
\begingroup%
  \makeatletter%
  \providecommand\color[2][]{%
    \errmessage{(Inkscape) Color is used for the text in Inkscape, but the package 'color.sty' is not loaded}%
    \renewcommand\color[2][]{}%
  }%
  \providecommand\transparent[1]{%
    \errmessage{(Inkscape) Transparency is used (non-zero) for the text in Inkscape, but the package 'transparent.sty' is not loaded}%
    \renewcommand\transparent[1]{}%
  }%
  \providecommand\rotatebox[2]{#2}%
  \newcommand*\fsize{\dimexpr\f@size pt\relax}%
  \newcommand*\lineheight[1]{\fontsize{\fsize}{#1\fsize}\selectfont}%
  \ifx\svgwidth\undefined%
    \setlength{\unitlength}{91.08703013bp}%
    \ifx\svgscale\undefined%
      \relax%
    \else%
      \setlength{\unitlength}{\unitlength * \real{\svgscale}}%
    \fi%
  \else%
    \setlength{\unitlength}{\svgwidth}%
  \fi%
  \global\let\svgwidth\undefined%
  \global\let\svgscale\undefined%
  \makeatother%
  \begin{picture}(1,0.80003428)%
    \lineheight{1}%
    \setlength\tabcolsep{0pt}%
    \put(0,0){\includegraphics[width=\unitlength,page=1]{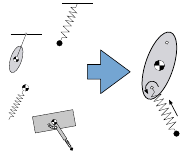}}%
    \put(-1.2919498,1.70509182){\color[rgb]{0,0,0}\makebox(0,0)[lt]{\begin{minipage}{5.91785364\unitlength}\raggedright \end{minipage}}}%

    \put(0.13690139,0.47280871){\color[rgb]{0,0,0}\makebox(0,0)[lt]{\lineheight{1.25}\smash{\begin{tabular}[t]{l}$\mathsf{VPP}$\end{tabular}}}}%
    \put(0.15650338,0.25626555){\color[rgb]{0,0,0}\makebox(0,0)[lt]{\lineheight{1.25}\smash{\begin{tabular}[t]{l}$\mathsf{SLIP}$\end{tabular}}}}%
    \put(0.41564508,0.65062481){\color[rgb]{0,0,0}\makebox(0,0)[lt]{\lineheight{1.25}\smash{\begin{tabular}[t]{l}$\mathsf{SLP}$\end{tabular}}}}%

    \put(0.42939336,0.10526519){\color[rgb]{0,0,0}\makebox(0,0)[lt]{\lineheight{1.25}\smash{\begin{tabular}[t]{l}$\mathsf{RHM}$\end{tabular}}}}%
  \end{picture}%
\endgroup%
 
	\caption[Unified Template Model]{Unified locomotion template model (ULT-model) that combines all locomotor subfunctions \cite{Seyfarth2015}: The repulsive leg function and postural stabilization during stance (inspired by the SLIP and VPP models) as well as the forward leg swing function (derived from RHM and SLP model) and leg retraction function (similar to SLP model) in the flight phase.}\label{fig:Concept}
\end{figure}
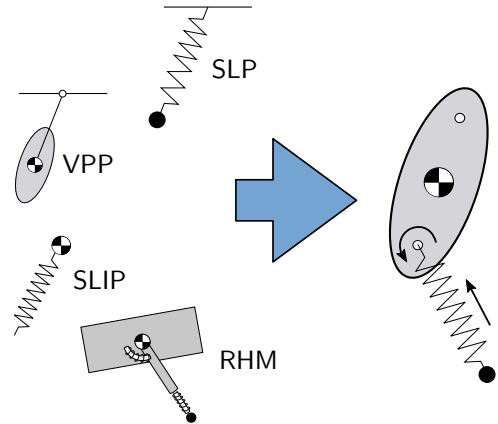
In typical SLIP models the legs are considered to be mass-less; therefore, the leg can be redirected without any effort during flight \cite{Blum2010}. In such models the only function of the swing phase is limited to optimally set the angle of attack for stable landing at the point of ground contact, which is associated with the net acceleration during the stance phase \cite{Raibert86}. In this case, the non-trivial effects of the leg forward swing on the trunk dynamics are ignored. For swing phase-specific models a rigid pendulum model is generally used, though spring-loaded pendulums have also been considered and shown to exert the expected torque on the hip \cite{Song2016}. One template model that is valid for stance and swing, which is typically used for robot hoppers (i.e. the robot hopper model, RHM), is made up of a rigid torso and leg with non-negligible mass \cite{thompson1990passive}. This model is typically used with the hip located at the CoM, which does not represent the trunk stabilization requirements of gait. A further limitation is that it does not characterize the leg retraction (and thus moment of inertia reduction) function during swing. \par

\begin{figure}
	\centering
	\vspace{0.25cm}
	\def\svgwidth{0.45\textwidth}
	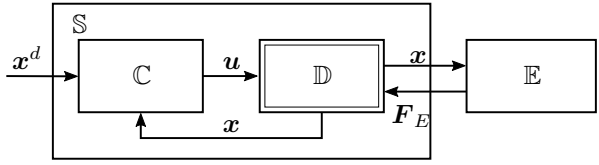 
	\caption[Blockdiagram]{The ideal ULT-model is composed of a compact system model $\mathbb{S}$ that is able to describe the system dynamics  $\mathbb{D}$ and interaction with the environment $\mathbb{E}$ during both stance and swing as well as a phase-independent controller $\mathbb{C}$. Here, the current and desired system state are denoted $\vect{x}$ and $\vect{x}^d$, respectively, and the control effort is represented by $\vect{u}$. The environment exerts friction and ground reaction forces $\vect{F}_E$, when the system is in contact.}
	\label{fig:Blockdiagram}
\end{figure}
In recent years, some approaches in bipedal locomotion control showed excellent results with very impressive locomotion capabilities \cite{ Englsberger, Caron, Guo2018}. However, they are computationally expensive. Generally, there is an intriguing discrepancy between the computational complexity required for biological locomotion versus robotic locomotion. This is clearly demonstrated by the fact that neurological feedback mechanisms are characterized by substantially lower signal frequency, lower control bandwidth and higher signal delay compared to robots \cite{Winter1998}, yet humans and animals are capable of highly dexterous, robust and stable locomotion. One possible explanation is that neuromechanical locomotion control is actually very simple because the underlying system behavior is inherently stable. This claim is supported in part by the well-known VPP-based template control models for stance \cite{Maus2008}, but it still remains to be shown for the full gait cycle. Furthermore, template-based locomotion models, which reflect the mechanisms by which the complexity of legged locomotion may be reduced, have proven difficult to implement in robotic systems and require complex controllers to compensate \cite{hutter2010slip, sreenath2011compliant, Hwang2018}. We hypothesize that the template models used to date do not capture the full essence of all relevant effects in the gait dynamics and thus make control more complex in a later stage. Therefore, the original question of designing the right template model remains. Accordingly, the fundamental research question behind the current work is whether we can find a unified and compact dynamic and control system (Fig. 2), which we call the unified locomotion template model (ULT-model) and is able to describe robustly stable legged locomotion. \\
\subsection{Contribution} \label{sec:contribution}
The specific focus and main contribution of this work is to develop a new unified and compact dynamic model with an upright trunk (in contrast to the RHM model) and non-negligible leg mass and moment of inertia ($\mathbb{D}$, Fig. 2), which is valid for both the stance and swing phase of gait. For feasibility testing purposes, a simple phase-based switching controller and hybrid contact model were applied to the new hypothesis as a first step\footnote{The unified control model is beyond the scope of the paper and will be pursued in subsequent steps.}. We aimed to show that all important subfunctions of locomotion could be enabled by augmenting a small number of model subcomponents to the previously described VPP-TSLIP model \cite{Maus2008} and that the resulting system can reach a stable and robust limit cycle with even a very simplistic controller. \par

For our proposed template model, we hypothesize the following:
\begin{itemize}
	\item[(H1)] The repulsive leg function during stance can be represented by a spring leg model. 
	\item[(H2)] By including a point mass at the foot and a simple resting length switch actuator along the axis of the leg spring, the retractive leg function of the swing phase is captured without compromising the stance behavior.
	\item[(H3)] Postural stabilization is achieved by applying a hip torque, which follows the VPP concept during stance and a velocity error-dependent set-point control law during swing.
	\item[(H4)] The same hip torque actuator and set-point controller generates the necessary forward leg swing to converge to the desired CoM velocity.
	\item[(H5)] With these model assumptions a stable and robust limit cycle can be found.
	\item[(H6)] Including the dynamic interaction between the torso and point foot in the model can actually lead to a more realistic representation of the overall gait behavior and is in accordance with findings from biomechanics.
\end{itemize}

We analyze the above hypotheses under the following assumptions:
\begin{itemize}
	\item[(A1)] Ground contact is modeled as a discontinuity.
	\item[(A2)] Control laws switch with gait phase.
\end{itemize}

\begin{figure}
	\centering
	\def\svgwidth{0.35\textwidth}
	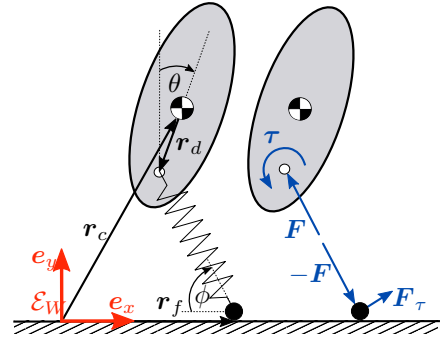 
	\vspace{-0.5cm}
	\caption[VPP-SLIP Model]{Coordinate definition and force diagram of the proposed dynamics model. The positions and orientations are defined with respect to a common world frame $\mathcal{E}_W$. A torque $\vect{\tau}$ that acts on the hip corresponds to a certain force $\vect{F}_{\tau}$ that acts on the foot and is perpendicular to the leg axis.}\label{fig:Model}
\end{figure}

\section{Methods} \label{sec:Methods}
\begin{figure}
	\centering
	\def\svgwidth{0.45\textwidth}
	\vspace{0.1cm}
	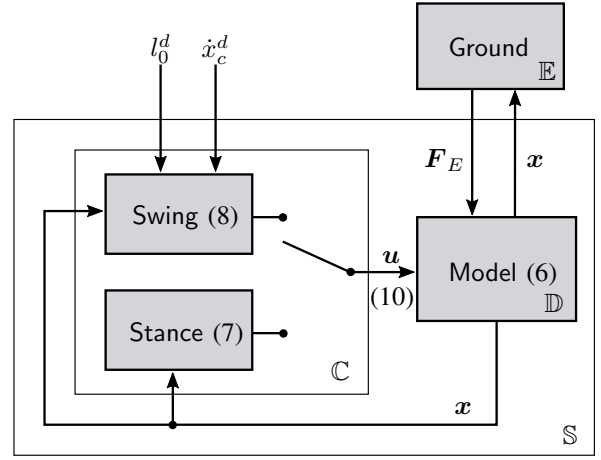 
	\caption[Blockdiagram]{System architecture. The two desired inputs that drive the system behavior are leg retraction, shaped by the desired resting length $l_0^d$, and system forward velocity, defined by $\dot{x}_c^d$. The ground reaction and friction forces exerted by the environment are denoted as $\vect{F}_E$. }\label{fig:Detail}
\end{figure}

In this section, we describe our model in detail (Section \ref{sec:modeling}) and introduce the control strategy (Section \ref{sec:control}). We also give an overview on proving orbital stability for the gait cycle in Section \ref{sec:stability}. A general overview of the system architecture is given in Fig. \ref{fig:Detail}.
\subsection{Modeling} \label{sec:modeling}
The proposed model consists of a rigid body and a point foot that are coupled by a spring leg attached to the hip. A schematic overview of the model depicted in Fig. \ref{fig:Model}. By means of actuators, a torque $\tau$ can be applied at the hip and the rest position of the leg spring can be changed to $l_0 + \xi$. We introduce the vector of control inputs $\vect{u} = \left[\tau, \xi \right]^{\mathsf{T}}$ as a compact notation. The Cartesian positions of the upper body's CoM and the point foot are defined as
\begin{align}
\vect{r}_c = \begin{bmatrix}
x_c \\ y_c
\end{bmatrix}\text{, }
\vect{r}_f = \begin{bmatrix}
x_f \\ y_f
\end{bmatrix}\text{, }
\end{align}
respectively. The vector from CoM to hip is denoted $\vect{r}_d$. The trunk inclination is denoted with $\theta$. With the upper body mass $m_c$, the foot mass $m_f$ and the trunk moment of inertia $J$, the equations of motion can be formulated as 
\begin{align}
m_c \ddot{\vect{r}}_c &= \vect{F}  + m_c \vect{g}  \label{eq:rc}\\
m_f \ddot{\vect{r}}_f &= -\vect{F} + \vect{F}_{\tau} + m_f \vect{g} + \vect{F}_E \label{eq:rf}\\
J \ddot{\theta} &= \tau  + d \left( F_{x} \cos(\theta) - F_{y} \sin(\theta) \right)\text{, } \label{eq:J}
\end{align} 
where 
\begin{equation}
\vect{F} = \begin{bmatrix}
F_{x} \\ F_{y}
\end{bmatrix} = k \left(\frac{l_{0} + \xi}{l} - 1\right) \vect{r}_l
\end{equation}
is the force exerted by the spring leg and the actuator, $\vect{F}_{\tau}$ is the force that results from the hip torque, \mbox{$\vect{g} = \left[0,  -g\right]^\mathsf{T}$} is the gravitational vector, $d = \norm{\vect{r}_d}$ is the distance from CoM to hip, $k$ is the spring constant, $l_0$ is the spring's rest length, $\vect{r}_l = \vect{r}_f -\vect{r}_c- \vect{r}_d$ is the vector pointing from foot to hip and $l = \norm{\vect{r}_l}$ is the current leg length. We combine all states in the joint system state vector $\vect{x} = \left[\vect{r}_c,\vect{r}_f,\theta, \dot{\vect{r}}_c,\dot{\vect{r}}_f,\dot{\theta} \right]^{\mathsf{T}}$. As a compact notation, we rewrite the equations of motion as \begin{equation}
\dot{\vect{x}} = \vect{f}(\vect{x})	= \begin{bmatrix}
\dot{\vect{r}}_c \\
\dot{\vect{r}}_f \\
\dot{\theta} \\
\frac{1}{m_c} \vect{F} + \vect{g} \\
\frac{1}{m_f} (-\vect{F} + \vect{F}_{\tau} + \vect{F}_E) + \vect{g}\\
\frac{1}{J} (\tau + d \left( F_{x} \cos(\theta) - F_{y} \sin(\theta) \right))
\end{bmatrix}
\text{.} \label{eq:vector_field}
\end{equation} 

The environmental force $\vect{F}_E = 0$ during flight. We do not need to explicitly compute $\vect{F}_E$ during stance, as it is assumed that the impact occurs instantaneously and that the friction forces are great enough that no slipping occurs. Accordingly, the foot is governed by $\ddot{\vect{r}}_f = \vect{0}$, $\dot{\vect{r}}_f = 0$ and ${\vect{r}}_f = const$ during stance.
Foot detachment happens as soon as the lift-off condition  $l > l_0$ is met. During flight, the system can move freely and evolves according to \eqref{eq:rc} - \eqref{eq:J}. It should be noted, since the foot sticks to the ground during the stance phase, the dynamics exactly match the VPP-TSLIP \cite{Maus2010}.

\subsection{Control} \label{sec:control}
Various control actions must be taken over the course of a gait cycle. We therefore first deal with stance phase control, before moving on to the swing phase.
\subsubsection{Stance Phase}
Maus et al. \cite{Maus2010} found that for postural stabilization during stance humans and animals mimick external support by redirecting the ground reaction forces into a metacenter above the CoM, the virtual pivot point (VPP). In this manner, a virtual pendulum is created from which the CoM is suspended. Disturbances lead to a deflection of the pendulum and corresponding swinging motions, which can be tolerated by the system. Thus, the system dynamics intrinsically stabilize the torso. This is radically different from common control approaches that rely on feedback control to stabilize an inverted pendulum, which is a naturally unstable system \cite{Kajita2001}, or from robotic system design approaches that involve mechanically placing the hip at the CoM (e.g. RHM model \cite{thompson1990passive}). \par
\begin{figure}
	\centering
	\def\svgwidth{0.35\textwidth}
\begingroup%
  \makeatletter%
  \providecommand\color[2][]{%
    \errmessage{(Inkscape) Color is used for the text in Inkscape, but the package 'color.sty' is not loaded}%
    \renewcommand\color[2][]{}%
  }%
  \providecommand\transparent[1]{%
    \errmessage{(Inkscape) Transparency is used (non-zero) for the text in Inkscape, but the package 'transparent.sty' is not loaded}%
    \renewcommand\transparent[1]{}%
  }%
  \providecommand\rotatebox[2]{#2}%
  \newcommand*\fsize{\dimexpr\f@size pt\relax}%
  \newcommand*\lineheight[1]{\fontsize{\fsize}{#1\fsize}\selectfont}%
  \ifx\svgwidth\undefined%
    \setlength{\unitlength}{103.60831914bp}%
    \ifx\svgscale\undefined%
      \relax%
    \else%
      \setlength{\unitlength}{\unitlength * \real{\svgscale}}%
    \fi%
  \else%
    \setlength{\unitlength}{\svgwidth}%
  \fi%
  \global\let\svgwidth\undefined%
  \global\let\svgscale\undefined%
  \makeatother%
  \begin{picture}(1,0.8138595)%
    \lineheight{1}%
    \setlength\tabcolsep{0pt}%
    \put(0,0){\includegraphics[width=\unitlength,page=1]{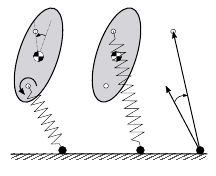}}%
    \put(0.78369592,0.24138282){\color[rgb]{0,0,0}\makebox(0,0)[lt]{\lineheight{1.25}\smash{\begin{tabular}[t]{l}$\vect{r}_l$\end{tabular}}}}%
    \put(0.03608907,0.66877019){\color[rgb]{0,0,0}\makebox(0,0)[lt]{\lineheight{1.25}\smash{\begin{tabular}[t]{l}$\mathsf{VPP}$\end{tabular}}}}%
    \put(0.17471447,0.60843914){\color[rgb]{0,0,0}\makebox(0,0)[lt]{\lineheight{1.25}\smash{\begin{tabular}[t]{l}$\delta$\end{tabular}}}}%
    \put(0.87069771,0.42407484){\color[rgb]{0,0,0}\makebox(0,0)[lt]{\lineheight{1.25}\smash{\begin{tabular}[t]{l}$\vect{r}_{\mathrm{VPP}}$\end{tabular}}}}%
    \put(0.31623225,0.50346807){\color[rgb]{0,0,0}\makebox(0,0)[lt]{\lineheight{1.25}\smash{\begin{tabular}[t]{l}\Large{$\equiv$}\end{tabular}}}}%
    \put(0.83531993,0.31675397){\color[rgb]{0,0,0}\makebox(0,0)[lt]{\lineheight{1.25}\smash{\begin{tabular}[t]{l}$\gamma$\end{tabular}}}}%
  \end{picture}%
\endgroup%

	\vspace{-0.5cm} 
	\caption[VPP-SLIP Model]{Virtual Pivot Point Concept. With an appropriate hip torque, the left model behaves exactly like the right one. The offset angle from the body axis to the VPP is denoted as $\delta$ and the angle between the vector pointing from foot to hip and from foot to VPP is denoted as $\gamma$.}\label{fig:VPP}
\end{figure}
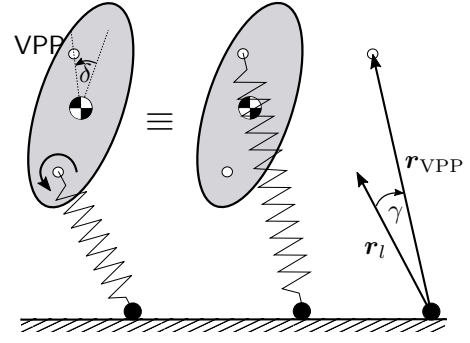

In order to implement the VPP controller on the system from section \ref{sec:modeling}, the force generated by the leg spring must be realigned. This can be done by introducing an appropriate tangential force at the foot using a respective hip torque. This concept is illustrated in Fig. \ref{fig:VPP}. 
As in \cite{Maus2008}, the hip motor torque is chosen to be 
\begin{equation}
\vect{\tau} = \norm{\vect{F}} \frac{\vect{r}_{\mathrm{VPP}} \times \vect{r}_l}{\vect{r}_{\mathrm{VPP}} \cdot \vect{r}_l} = \begin{bmatrix}
0 \\
0 \\
\norm{\vect{F}}  l \tan(\gamma)
\end{bmatrix}\text{,} \label{eq:VPP}
\end{equation}
where $\vect{r}_{\mathrm{VPP}}$ is the vector pointing from the point foot to the VPP. During stance phase, the value for the leg length actuator is set to $\xi = 0$. By proper choice of initial values and model parameters, the spring will compress, store and release energy to propel the model forward. Upper body stabilization is solely achieved by the virtual pendulum mechanics. Since the foot mass is in contact with the ground, the stance phase is the same as for the example described in \cite{Sharbafi2012}.
\subsubsection{Swing Phase}
The swing phase has two main functions: leg retraction and foot placement. Due to the conservation of angular momentum during flight, maintaining the upper-body orientation while swinging the leg forward is not possible. This effect can be alleviated by bringing the foot closer to the COM, thus reducing the leg moment of inertia. The foot must also be correctly positioned during the swing phase to ensure a favorable system configuration for the stance phase at the moment of touch-down. Here, instead of complex trajectory planning, the swing-phase behavior is simply adjusted by two set points. For leg retraction, the resting length of the spring is changed to a new value with some fixed $\xi = l_0^d - l_0 $. For the forward swinging motion, the leg angle of attack is controlled using a simple torque control law 
\begin{equation}
\tau = c(\varphi_d - \varphi) - b\dot{\varphi}, \label{eq:swing}
\end{equation}
where $c$ and $b$ are the angular stiffness and damping values, respectively, $\varphi$ and $\varphi_d$ are the actual and desired angle of attack and $\dot{\varphi}$ is the angular velocity. \\
As Raibert showed in \cite{Raibert86}, the orientation of the leg at touch-down determines the acceleration behavior in the subsequent stance phase, with the neutral point defined as the angle that yields zero net acceleration for a given forward velocity. This information has been proven to be advantageous for defining the angle of attack that is needed for each step so that the system can recover from disturbances and return to a nominal forward speed. Calculating the neutral point requires information on the exact length of the swing phase, which might not be a priori known, since the system might be subject to a sudden disturbance. We therefore opt for a simple feedback control law that is evaluated once per gait cycle after touch-down:
\begin{equation}
\varphi_d^{\mathrm{next}} = \varphi_0 + K (\dot{x}_c^d - \dot{x}_c). \label{eq:phi_law}
\end{equation}
Here, $\varphi_d^{\mathrm{next}}$ is the angle of attack for the next swing phase, $\varphi_0$ is the initial angle of attack, $K$ is a feedback gain and $\dot{x}_c^d$ is the desired forward velocity of the CoM. If the current forward velocity deviates from the desired one, the angle of attack is automatically adjusted. If $K$ is properly tuned, the system converges to an angle of attack that corresponds to a neutral point after a few gait cycles. \\
With the individual controllers for stance and swing described, we summarize the designed control action as
\begin{equation}
\vect{u}= \begin{cases}
\begin{bmatrix} 
\norm{\vect{F}}  l \tan(\gamma) \\
0 \end{bmatrix}, & \text{during stance},\\[10pt]
\begin{bmatrix} 
c(\varphi_d - \varphi) - b\dot{\varphi}\\
l_0^d - l_0 \end{bmatrix}, & \text{during swing}.
\end{cases} \label{eq:control}
\end{equation}

\subsection{Stability} \label{sec:stability}
When looking at solutions of the vector field \eqref{eq:vector_field}, we are interested in mainly two properties. The first one is periodicity; meaning the system returns to the same configuration after each period, i.e. gait cycle. The second property is robustness with respect to disturbances: If a deviation from the nominal gait cycle leads to failure, the system might not be orbitally stable \cite{Sobotka2007}. In this section, we introduce the necessary tools, in order to evaluate orbital stability. The following definitions are taken from \cite{Parker2012} and \cite{Sobotka2007}.
\begin{definition}
	A solution $\vect{\phi}_t(\vect{x}^*)$ of $\vect{f}(\vect{x})$ with the initial configuration $\vect{x}^*$ is periodic, if 
	\begin{equation}
	\vect{\phi}_t(\vect{x}^*) = \vect{\phi}_{t+T}(\vect{x}^*),
	\end{equation}
	for all $t$ and $T>0$. Here, any $\vect{x}^*$ that lies on the periodic solution can serve as a starting value.  
\end{definition}
If the periodic solution is isolated, i.e. there exist a neighbourhood around it in which no other periodic solutions exists, then it is called a limit cycle. The curve traced by the limit cycle is denoted as $\Gamma$. A common practice to analyze dynamical systems such as the system at hand is the method of Poincar\'{e} maps. Instead of viewing the solution of an $n$th-order dynamical system, a $(n-1)$-order discrete system termed Poincar\'{e} map is introduced. 
\begin{definition}
	Let $\Gamma$ be a curve corresponding to a limit cycle. Consider also an $(n-1)$-dimensional cross-section $\Sigma$ that is transversal to $\Gamma$. Let $\vect{x}^*$ be the point on $\Gamma$ that intersects $\Sigma$. We now define the Poincar\'{e} map {$P: U \rightarrow \Sigma$} in a neighbourhood $U \subset \Sigma$ of $\vect{x}^*$ as 
	\begin{equation}
	\vect{P}(\vect{x}) := \vect{\phi}_{t_0} (\vect{x}) \text{,}
	\end{equation}
	where $t_0$ is the first time the flow hits the cross-section after starting at a value $\vect{x} \subset U$. By definition, the Poincar\'{e} map has a fixed point at $\vect{x}^*$ and $\vect{P}(\vect{x}^*) = \vect{x}^*$ holds. 
\end{definition}


With the above definition, a limit cycle corresponds to a fixed point of the Poincar\'{e} map. Thus, by proving the stability of the fixed point, we can simultaneously prove stability of the limit cycle. The stability of the fixed point can be assessed by evaluating the Poincar\'{e} map linearization about the fixed point
\begin{equation}
\vect{DP}(\vect{x}) = \frac{\partial \vect{P}(\vect{x})}{\partial \vect{x}}\Bigg|_{\vect{x} =\vect{x}^* }.
\end{equation} 
A limit cycle is periodically stable, if all the eigenvalues of the linearized  Poincar\'{e} map (the so-called Floquet multipliers) lie within the unit cycle \cite{Guckenheimer1984} (Lyapunov's indirect method). One possible choice of the Poincar\'{e} map is the apex return map which maps the height of two consecutive apex states.  Since there is no analytical expression of this map in our case, it has to be evaluated numerically. This can be done by perturbation of the initial conditions. The linearized  Poincar\'{e} map can be then computed by central differences
\begin{equation}
\vect{DP}(\vect{x}) = \frac{\vect{P}(\vect{x} + \epsilon \vect{\chi}) - \vect{P}(\vect{x} -\epsilon \vect{\chi})}{2 \epsilon} \text{,}
\end{equation}
where $\epsilon$ is a small positive scalar value and $\vect{\chi}$ is the direction of the perturbation.

\begin{table}
	\vspace{0.3cm}
	\label{tbl:values}
	\centering
	\caption{Model and control parameter values. The mechanical parameters were chosen similar to \cite{Sharbafi14}.}
	
	\begin{tabularx}{0.4\textwidth}{c|l|c|l}
		
		Parameter & Value & Parameter & Value\\
		\hline  
		& & & \\
		$m_c$ & $80.0$  $\mathrm{kg}$ & $l_0$  & $1.0$  $ \mathrm{m}$ \Tstrut\\ 
		$m_f$ & $3.4$  $\mathrm{kg}$ & $k$  & $21$  $ \mathrm{kN}$  $\mathrm{m}^{-1}$\\
		$J$  & $5.0$  $ \mathrm{kg} \mathrm{m}^2$ & $c$  & $1.9$ $\mathrm{kNm}$ $ \mathrm{rad}^{-1}$\\
		$d$  & $0.1$  $ \mathrm{m} $ & $b$  & $80.37$ $\mathrm{Nm s}$ $\mathrm{rad}^{-1}$\\
		$\varphi_0$ & $70.0$  $\deg$ $ \mathrm{m} $ & $K$ & $0.15$\\
		$d_{\mathrm{VPP}}$  & $0.25$  & & \\
		$\delta$  & $0.0$  $ \mathrm{rad}$ & & \\
	\end{tabularx} \\
	\vspace{-0.25cm}
	
\end{table}

\section{Results}\label{sec:results}
In this section, we report the simulation results for two exemplary limit cycles. The simulation is carried out using an ODE solver with rootfinding capability provided by the SUNDIALS framework (SUite of Nonlinear and DIfferential/ALgebraic Equation Solvers) \cite{Sundials}. The absolute and relative tolerances are set to $10^{-10}$ and $10^{-9}$, respectively. For the system, we used the model and control parameters reported in Table I. \par
\begin{figure}
	\centering
	\scriptsize 
	\begin{minipage}{\textwidth}
		\begin{minipage}{0.4\textwidth}
			\begin{tabularx}{0.5\textwidth}{c|c}
				'$\Diamond$' & 'X'\\
				\hline
				$\begin{bmatrix} 0.0 \Tstrut\\
				1.0354\\
				-0.3816\\
				0.0071\\
				0.4264\\
				3.9601\\
				0.0\\
				4.238\\
				2.7802\\
				-0.4410
				\end{bmatrix}$ &$\begin{bmatrix} 0.0\\
				1.0319\\
				-0.893\\
				0.0075\\
				0.453\\
				3.994\\
				0.0\\
				5.723\\
				2.100\\
				-0.286
				\end{bmatrix}$ 
			\end{tabularx} 
		\end{minipage}
		\begin{minipage}{\textwidth}
			\hspace{-3cm} 
			\raisebox{-1.75cm}{\includegraphics[width = 0.22\textwidth]{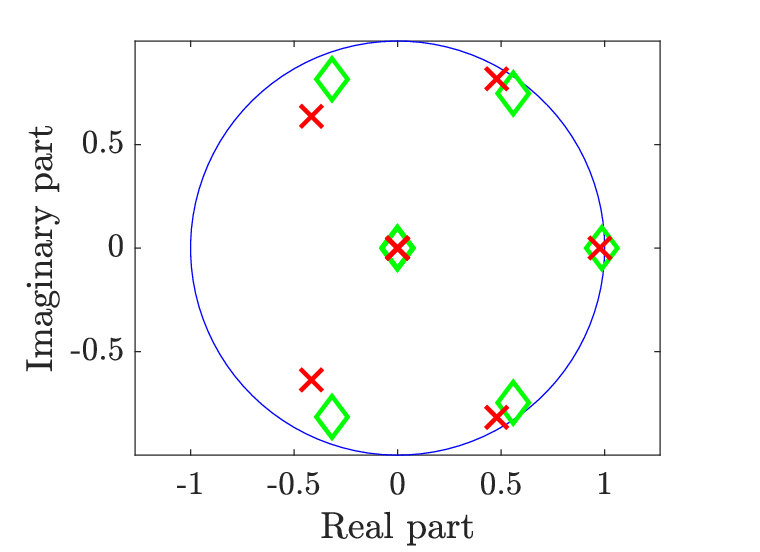}}
		\end{minipage}
	\end{minipage}
	\caption{ Fixed points and corresponding eigenvalues. The first and second fixed point correspond to the system marked with green '$\Diamond$' and red 'X' in Fig. \ref{fig:sweep}, respectively.} \label{eigen}
\end{figure}
\begin{figure}
	\hspace*{-1cm}	\centering
	\includegraphics[width=0.35\textwidth]{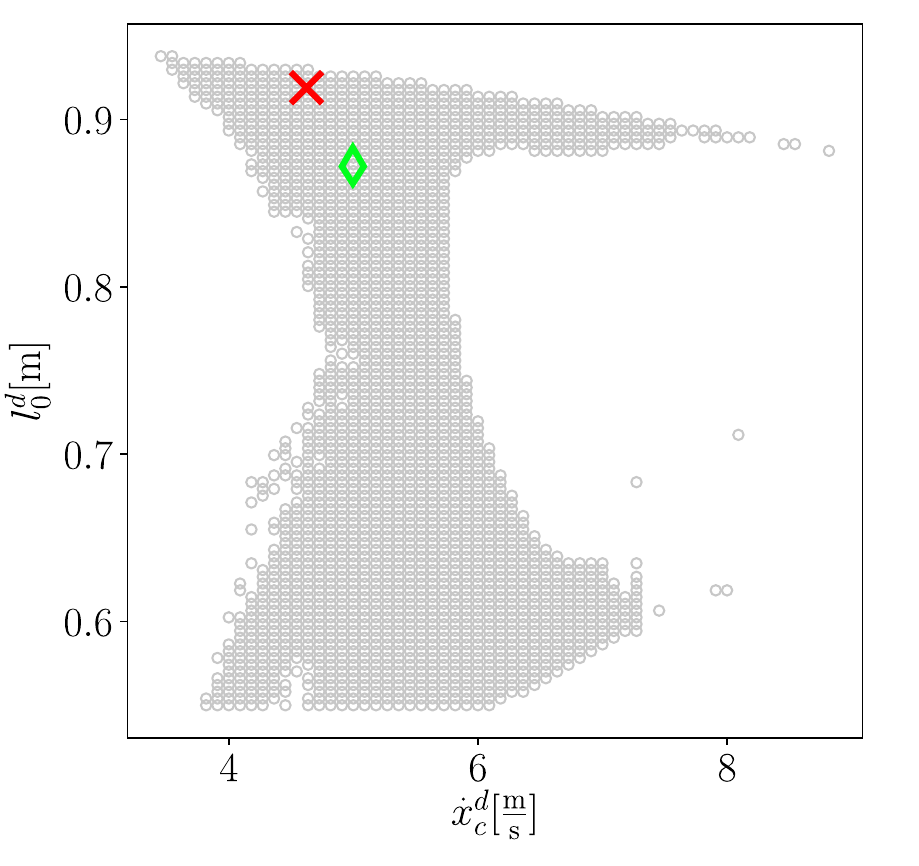}
	\vspace{-0.5cm}
	\caption[Simulation]{Candidates for stable limit cycles in the parameter space of the desired CoM forward velocity $\dot{x}_c^d$ and leg retraction $l_0^d$. The limit cycles corresponding to $(\dot{x}_c^d, l_0^d) = (5.0, 0.087)$ (marked with a green '$\Diamond$') and $(\dot{x}_c^d, l_0^d) = (4.6, 0.091)$ (marked with a red 'X') are chosen as exemplary limit cycles for our analysis.} \label{fig:sweep}
\end{figure}
\begin{figure}
	\centering
	\includegraphics[width=0.5\textwidth]{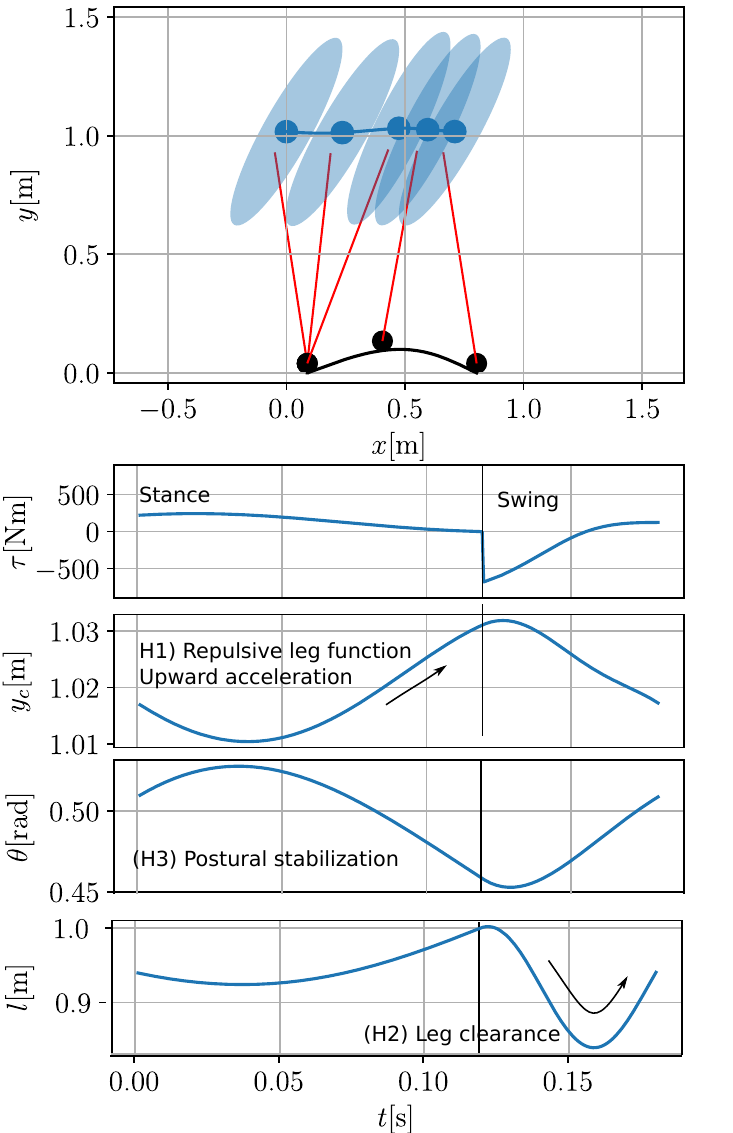}
	\vspace{-0.25cm}
	\caption[Simulation]{(Above) A sequence of position captures demonstrates the manner in which the torso and leg angles develop over a full gait cycle (stance and swing). (Below) The hip joint torque is positive (extension moment) during stance and cycles from flexion to extension during swing. The data for torso height, leg length and trunk inclination is in accordance with our expectations.} \label{fig:Simulation}
\end{figure}

In order to set the stage for further analysis, we search the parameter space of $\dot{x}_c^d$ and $l_0^d$ to identify candidates for stable limit cycles. For each parameter combination in the search space, stable limit cycle candidates are found by checking whether the model is still progressing forward or has fallen after one hundred steps (i.e steps to fall method \cite{Seyfarth2002}). 
The region of candidates is depicted in Fig. \ref{fig:sweep}. For the two candidates marked in the figure, stability is checked by using the linearized Poincar\'{e} map. 
The fixed points and eigenvalues that correspond to the respective limit cycles are reported in Fig. \ref{eigen}. Because the eigenvalues for both candidates lie within the unit cycle, both candidates proved to be indeed stable. In order to further check the robustness of the two selected limit cycles, a perturbation to the initial height of $-7.5\%$ was applied. As can be seen from {Fig. \ref{fig:limit_cycles}}, the system corresponding to the parameters in the center of the identified region (Fig. \ref{fig:sweep}, green '$\Diamond$') was able to converge back to the limit cycle despite perturbation, indicating robustness. The system corresponding to the parameters at the edge of the identified region (Fig. \ref{fig:sweep}, red 'X'), on the other hand, diverged from the limit cycle, despite having roughly comparable eigenvalues. A full gait cycle with the more robust system is depicted in Fig. \ref{fig:Simulation}. 

%
%

\section{Discussion}\label{sec:discussion}
\begin{figure}
	\vspace{0.25cm}
	\centering
	\def\svgwidth{\columnwidth}
	\includegraphics[width=\columnwidth]{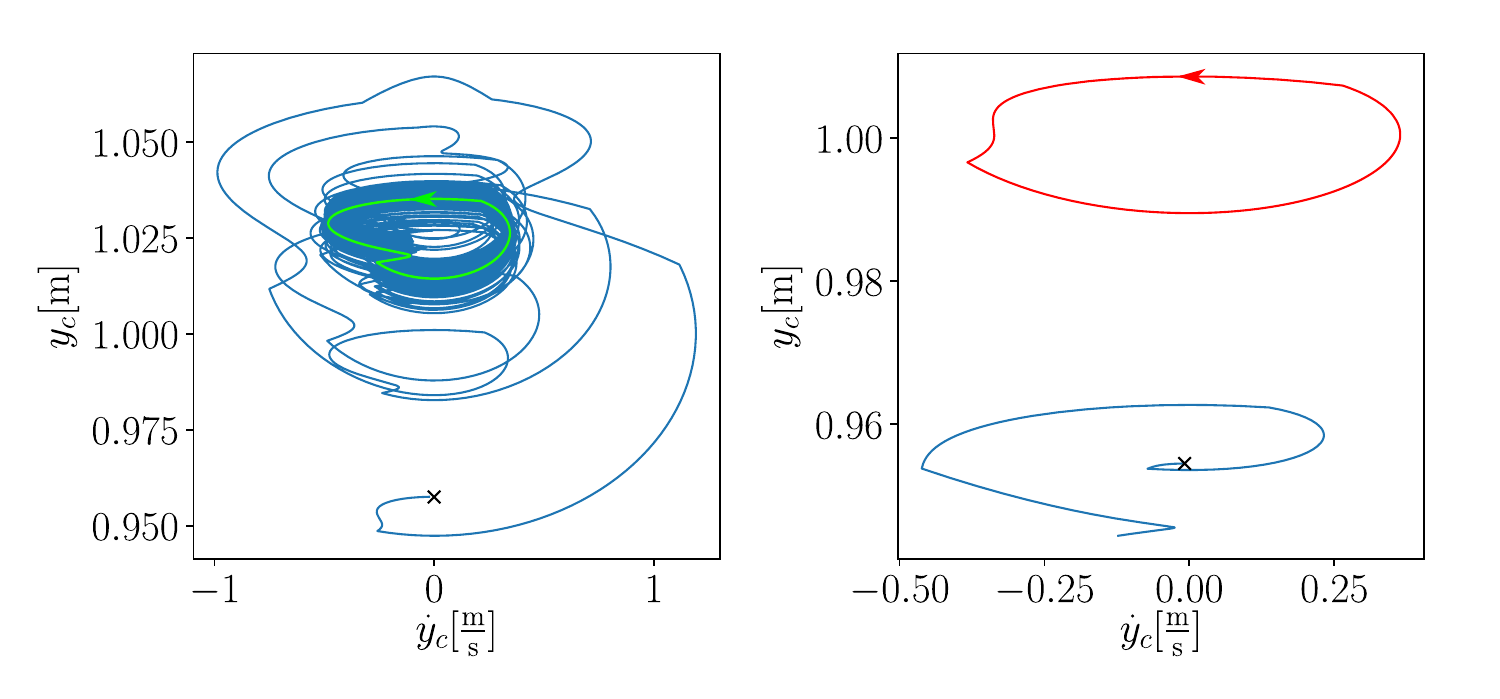}
\vspace{-0.75cm}
	\caption[Perturb]{Phase portraits for two limit cycles with perturbed initial condition corresponding to the green '$\Diamond$' (left) and red 'X' (right) from Fig. \ref{fig:sweep}. The left example converges to the limit cycle after perturbation while the system on the right becomes unstable and diverges from the limit cycle.} \label{fig:limit_cycles}
\end{figure}

\begin{figure}
	\centering
	\def\svgwidth{\columnwidth}
	\includegraphics[width=\columnwidth]{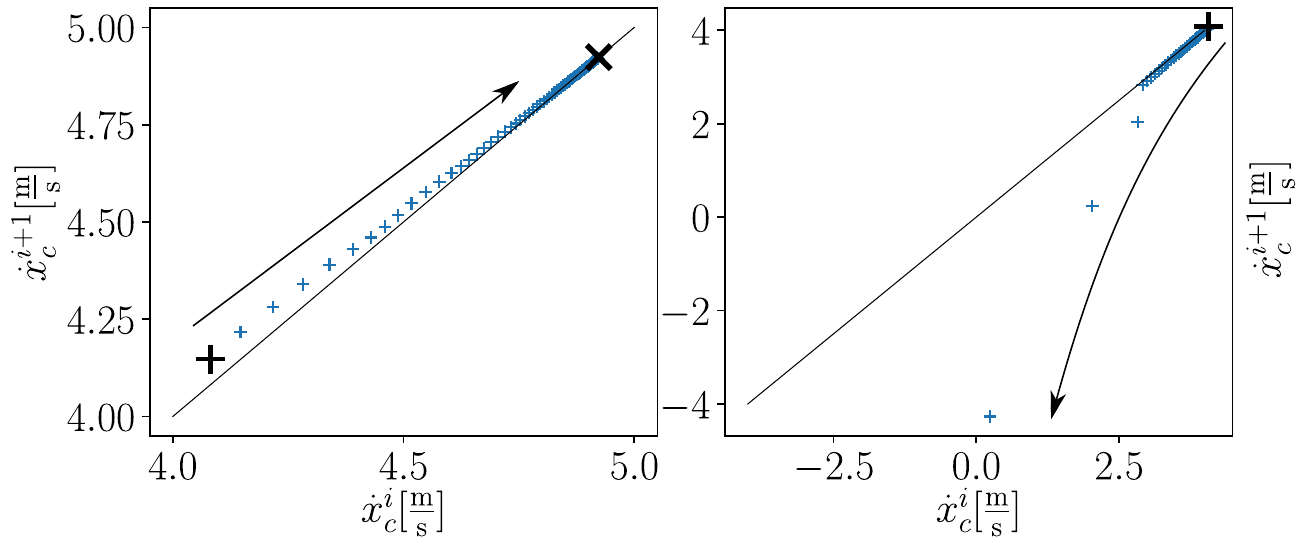}
	\vspace{-0.25cm}
	\caption[Perturb]{Poincar\'{e} map for the CoM velocity without a controller. (Left) If $\varphi^d$ is continuously adapted according to the control law \eqref{eq:phi_law}, the system converges to a fixed point. (Right) In case of a constant $\varphi^d$, the system decelerates and becomes unstable. The initial value is marked with '+' and the fixed point (which only exists for the left case) is marked with 'x'.} \label{fig:poincare}
\end{figure}

In this work, we have taken a significant first step toward developing a unified template model for an upright-trunk forward hopping system by defining a new non-discretized and compact system model. This model inherently describes both the stance and swing phase with minimal but meaningful changes to the previously defined VPP-TSLIP model. The chosen model elements generated the desired behavior and a stable limit cycle was found for the applied switch controller. Each of our initial hypotheses was confirmed as follows:
\begin{itemize}
	\item[(H1)] The leg spring length decreased during the first half of stance and increased during the second half while accelerating the torso upward (Fig. \ref{fig:Simulation}), demonstrating the expected repulsive function. 
	\item[(H2)] The leg spring length was observed to decrease during the first half of the flight phase and increased during the second half, as expected (Fig. \ref{fig:Simulation}). 
	\item[(H3)] Postural stability was achieved throughout the gait cycle, as hypothesized (Fig. \ref{fig:Simulation}, Fig. \ref{fig:limit_cycles}).
	\item[(H4)] By actuating the forward leg swing according to the error between the desired and actual forward CoM velocity, the model was able to converge to the desired velocity after perturbation (Fig. \ref{fig:poincare}, left). In contrast, a fixed angle of attack led to unstable behavior (Fig. \ref{fig:poincare}, right), confirming our hypothesis.
	\item[(H5)] Stable regions were found for the desired CoM velocity and leg retraction (Fig. \ref{fig:sweep}) and points within the regions were shown to be robust to perturbations (Fig. \ref{fig:limit_cycles}).
	\item[(H6)] As expected, the dynamic behavior of the model mimicked certain biological kinematics and kinetics, which is outlined in more detail below.
\end{itemize}
One interesting observation regarding the model's dynamic behavior is that the average torso inclination angle throughout the gait cycle is slightly forward. During stance, the torso is verticalized, i.e. it exhibits a negative angular velocity. Due to conservation of angular momentum, the torso would continue to rotate backwards after lift-off in the absence of external influence. However, the forward leg swing induces an angular acceleration in the opposite direction, which generates a positive angular velocity in the torso. These findings are similar to observations of human data during running \cite{Riley2008}. This leads to the conclusion that the stance and swing phases must complement each other. A torso angular velocity that would have led to failure in previous models, in which the swing phase dynamics were ignored, is now compensated by the leg swing. The swing phase dynamics may also play a role in the upper body orientation converging to a slightly inclined angle. \par
The hip torque observed in the stable limit cycle is characterized by several interesting features as well. During the stance phase the torque is positive with a trend toward zero, corresponding to hip extension. This differs from two-legged running results in template models \cite{Maus2008Advances} and in humans \cite{Fukuchi2017}, but aligns with observations of forward jumping humans \cite{Suzuki2018}. There is a large jump in the hip torque at the transition from stance to swing, which does not reflect physiological trends and can be explained by the discretized control strategy applied here. A unified control strategy, which is planned in future work, is expected to smoothen this transition.

\vspace*{-0.25cm}

\section{Conclusion} \label{sec:conclusion}
In this paper, we presented the first step towards developing a unified locomotion template model (ULT-model) for the complete gait cycle dynamics of an upright-trunk hopping system. Using a simple switching control law, stable limit cycles emerge for various desired speeds and leg clearances. This unified approach yielded simulation results that more closely match human and animal motion data. Next steps include a uniform control strategy across the gait cycle and a more suitable contact model. It is expected that the resulting unified template model has significant potential by providing closed-loop dynamics over the full gait cycle and thus enabling true, non-discretized anchor matching. Also, the resulting system may enable better understanding of neuromuscular function (in particular during phase transitions) for upright-trunk locomotion as well as the development of legged robot controls that are able to dynamically react to disturbances without requiring replanning.
\section*{Acknowledgments}
We gratefully acknowledge the funding support of Microsoft Germany, the Alfried Krupp von Bohlen and Halbach Foundation, and the the EU Horizon 2020 research and innovation program under grant no. 101017274 (Darko).

\addtolength{\textheight}{-8cm}
\bibliographystyle{ieeetr}
\bibliography{references}

\end{document}